\documentclass[11pt,a4paper]{article}
\pdfoutput=1
\usepackage[nohyperref]{acl2019}
\usepackage{times}
\usepackage{latexsym}

\usepackage{url}

\aclfinalcopy 
\def\aclpaperid{1479} 

\usepackage{times}
\usepackage{latexsym}

\usepackage{url}

\usepackage{graphicx}

\usepackage{verbatim}
\usepackage{amsmath}
\usepackage{amsfonts}
\usepackage{amssymb}
\usepackage{bm}
\usepackage{bbm}
\usepackage{varwidth}
\usepackage{todonotes}
\usepackage[noend]{algpseudocode}
\usepackage{algorithm}
\usepackage{array}
\usepackage{multirow}
\usepackage{varwidth}
\usepackage[position=top]{subfig}

\usepackage{pgfplots}
\pgfplotsset{compat=1.14}

\usepackage{tikz}
\usepackage{tikz-dependency}
\usetikzlibrary{shapes,arrows,positioning,calc,patterns,fit,backgrounds}

\usepackage{amsmath,amsfonts,bm}

\def\eqref#1{equation~\ref{#1}}

\def\1{\bm{1}}

\def\vb{{\bm{b}}}

\def\ve{{\bm{e}}}

\def\vm{{\bm{m}}}

\def\vs{{\bm{s}}}

\def\vv{{\bm{v}}}

\def\vz{{\bm{z}}}

\def\evb{{b}}

\def\evs{{s}}

\def\mE{{\bm{E}}}

\def\mG{{\bm{G}}}

\def\mT{{\bm{T}}}

\def\mW{{\bm{W}}}

\DeclareMathAlphabet{\mathsfit}{\encodingdefault}{\sfdefault}{m}{sl}
\SetMathAlphabet{\mathsfit}{bold}{\encodingdefault}{\sfdefault}{bx}{n}

\def\sD{{\mathbb{D}}}

\def\emG{{G}}

\def\emT{{T}}
\def\emU{{U}}

\def\emW{{W}}

\newcommand{\E}{\mathbb{E}}

\newcommand{\R}{\mathbb{R}}

\newcommand{\softmax}{\mathrm{softmax}}

\DeclareMathOperator*{\argmax}{arg\,max}
\DeclareMathOperator*{\argmin}{arg\,min}

\newcommand{\relu}{\mathrm{ReLU}}
\newcommand{\mlphead}{\mathrm{MLP}^{\text{head}}}
\newcommand{\mlpmod}{\mathrm{MLP}^{\text{mod}}}

\DeclareMathOperator*{\onehotargmax}{one-hot-argmax}

\title{
    Learning Latent Trees with Stochastic Perturbations \\ and Differentiable Dynamic Programming 
}

\author{%
\hspace{-0.7cm}\begin{tabular}{>{\raggedleft}p{0.5\textwidth}>{\raggedright\arraybackslash}p{0.5\textwidth}}
Caio Corro~~~~ & ~~~~Ivan Titov
\end{tabular}\\
\hspace{-0.7cm}\begin{tabular}{>{\raggedleft}p{0.5\textwidth}>{\raggedright\arraybackslash}p{0.5\textwidth}}
\multicolumn{2}{c}{ILCC, School of Informatics, University of Edinburgh} \\
\multicolumn{2}{c}{ILLC, University of Amsterdam} \\
{\tt c.f.corro@uva.nl}~~~~&~~~~{\tt ititov@inf.ed.ac.uk}
\end{tabular}}

\date{}

\begin{document}
\maketitle
\begin{abstract}

We treat projective dependency trees as latent variables in our probabilistic model and induce them in such a way as to be beneficial for a downstream task, without relying on any direct tree supervision.
Our approach relies on Gumbel perturbations and differentiable dynamic programming. 
Unlike previous approaches to latent tree learning, we stochastically sample global structures and our parser is fully differentiable.
We illustrate its effectiveness on sentiment analysis and natural language inference tasks.
We also study its properties on a synthetic structure induction task.
Ablation studies emphasize the importance of both stochasticity and constraining latent structures to be projective trees.

\end{abstract}

\section{Introduction}

Discrete structures are ubiquitous in the study of natural languages, for example in morphology, syntax and discourse analysis.
In natural language processing, they are often used to inject linguistic prior knowledge into statistical models.
For examples, syntactic  structures have been shown beneficial in question answering~\cite{cui2005question}, sentiment analysis~\cite{socher2013sst},  machine translation~\cite{bastings2017translation} and relation extraction~\cite{liu2015dependency}, among others. However, linguistic tools producing these structured representations (e.g., syntactic parsers) are not available for many languages and not robust when applied outside of the domain they were trained on ~\cite{petrov2010parsing,foster2011hardtoparse}.
Moreover, linguistic structures do not always seem suitable in downstream applications, with simpler alternatives sometimes yielding better performance~\cite{wang2018tree}.

Indeed, a parallel line of work focused on inducing task-specific structured representations of language~\cite{naradowsky2012latent,yogatama2016reinforce,kim2017structuredatt,liu2018structuredatt,niculae2018dyncg}.
In these approaches, no syntactic or semantic annotation is needed for training: representation is induced from scratch in an end-to-end fashion, in such a way as to benefit a given downstream task.
In other words, these approaches provide an inductive bias specifying that (hierarchical) structures 
are appropriate for representing a natural language, but do not make any further assumptions regarding what the structures represent. Structures induced in this way, though useful for the task, tend not to resemble any accepted syntactic or semantic formalisms~\cite{williams2018latent}. Our approach falls under this category. 

In our method,
projective dependency trees (see Figure~\ref{tab:examples} for examples) are treated as latent variables within a probabilistic model. 
We rely on differentiable dynamic programming \cite{mensch2018differentiable} which allows for efficient sampling  of dependency trees \cite{corro2018differentiable}.
Intuitively, sampling a tree involves stochastically perturbing dependency weights and then running a relaxed form of the Eisner dynamic programming algortihm~\cite{eisner1996dep}.
A sampled tree (or its continuous relaxation) can then be straightforwardly integrated in a neural sentence encoder for a target task using graph convolutional  networks~\cite[GCNs,][]{kipf2016semi}. The entire model, including the parser and GCN parameters, are estimated jointly while minimizing the loss for the target task.

What distinguishes us from  previous work is that we stochastically sample global structures  and do it in a differentiable fashion.
For example, the structured attention method \cite{kim2017structuredatt,liu2018structuredatt} does not sample entire trees but rather computes arc marginals, and hence does not faithfully represent higher-order statistics.
Much of other previous work relies either on reinforcement learning \cite{yogatama2016reinforce,nangia2018listops,williams2018latent} or does not treat the latent structure as a random variable~\cite{head2018sigpot}.
\citet{niculae2018dyncg} marginalizes over latent structures, however, 
this necessitates strong sparsity assumptions on the posterior distributions which may inject undesirable biases in the model.
Overall, differential dynamic programming has not been actively studied in the task-specific tree induction context.
Most previous work also focused on constituent trees rather than dependency ones. 

We study properties of our approach on a synthetic structure induction task and  experiment on  sentiment classification  \cite{socher2013sst} and  natural language inference \cite{bowman2015snli}.  
Our experiments confirm that the structural bias encoded in our approach is beneficial.
For example, our approach achieves a 4.9\% improvement on multi-genre natural language inference (MultiNLI) over a structure-agnostic baseline.
We show that stochastisticity and  higher-order statistics given by the global inference are both important.
In ablation experiments,  we also observe that forcing the structures to be projective dependency trees rather than permitting any general graphs yields substantial improvements without sacrificing execution time.
This confirms that our inductive bias is useful, at least in the context of the considered downstream applications.%
\footnote{The Dynet code for differentiable dynamic programming is available at \url{https://github.com/FilippoC/diffdp}.}
Our main contributions can be summarized as follows:
\begin{enumerate}
\vspace{-1ex}
    \item we show that a latent tree model can be estimated by drawing global approximate samples via Gumbel perturbation and differentiable dynamic programming;
\vspace{-1ex}    
    \item we demonstrate that constraining the structures to be projective dependency trees  is beneficial;
\vspace{-1ex}
    \item we  show the effectiveness of our approach on two standard tasks used in latent structure modelling and on a synthetic dataset.
\end{enumerate}

\section{Background}

In this section, we describe the dependency parsing problem and GCNs which we use
to incorporate latent structures into models for downstream tasks.

\subsection{Dependency Parsing}

Dependency trees represent bi-lexical relations between words.
They are commonly represented as directed graphs with  vertices
and arcs corresponding to words and relations, respectively.

Let $x = x_0 \ldots x_n$ be an input sentence with $n$ words where $x_0$ is a special root token.
We describe a dependency tree of $x$ with its adjacency matrix $\mT \in \{0, 1\}^{n \times n}$ where $\emT_{h, m} = 1$ iff there is a relation from head word $x_h$ to modifier word $x_m$.
We write $\mathcal{T}(x)$ to denote the set of trees compatible with sentence $x$.

We focus on projective dependency trees.
A dependency tree $\mT$ is projective iff for every arc $\emT_{h, m} = 1$, there is a path with arcs in $\mT$ from $x_h$ to each word $x_i$ such that $h < i < m$ or $m < i < h$. Intuitively,
a tree is projective as long as it can be drawn above the words in such way that arcs do not cross each other (see Figure~\ref{tab:examples}).
Similarly to phrase-structure trees, projective dependency trees
implicitly encode hierarchical decomposition of a sentence into  spans (`phrases').
Forcing trees to be projective may be desirable as even flat span structures can be beneficial in applications (e.g., encoding multi-word expressions).
Note that actual syntactic trees are also, to a large degree, projective, especially for such morphologically impoverished languages as English.
Moreover, restricting the space of the latent structures is important to ease their estimation. 
For all these reasons, in this work we focus on projective dependency trees.

In practice, a dependency parser is given a sentence $x$ and predicts a dependency tree $\mT \in \mathcal{T}(x)$ for this input.
To this end, the first step is to compute a matrix $\mW \in \R^{n \times n}$ that scores each dependency.
In this paper, we rely on a deep dotted attention network.
Let $\ve_0 \ldots \ve_n$ be embeddings associated with each word of the sentence.%
\footnote{
    The embeddings can be context-sensitive, e.g., an     RNN state.
}
We follow \citet{parikh2016decomposable} and compute the score for each head-modifier pair $(x_h,x_m)$ as follows:
\begin{align}
\label{eq:parse-pred}
    \emW_{h, m}\! =\! \mlphead(\ve_h)\!^\top\! \mlpmod(\ve_m\!)\!+\!b_{h{\text -}m},
\end{align}
where $\mlphead$ and $\mlpmod$ are multilayer perceptrons, and $b_{h{\text -}m}$ is a distance-dependent bias, letting the model encode preference for long or short-distance dependencies.
The conditional probability of a tree $p_\theta(\mT | x)$ is defined by a log-linear model:
\begin{align*}
    p_\theta(\mT |  x) = \frac{\exp(\sum_{h, m} \emW_{h, m} \emT_{h, m})}{\sum_{T' \in \mathcal{T}(x)} \exp(\sum_{h, m} \emW_{h, m} \emT'_{h, m})}.
\end{align*}
When tree annotation is provided in data $\sD$, networks parameters $\theta$  are learned by maximizing the log-likelihood of annotated trees \cite{lafferty2001crf}.

The highest scoring  dependency tree can be produced
by solving the following mathematical program:
\begin{align}   
    \bm T = \argmax_{\bm T \in \mathcal{T}(\bm x)} \quad \sum_{h, m} \emW_{h, m} \emT_{h, m}. \label{eq:map}
\end{align}
If $\mathcal{T}(\bm x)$ is restricted to be the set of projective dependency trees,
this can be done efficiently in $O(n^3)$ using the dynamic programming algorithm of \citet{eisner1996dep}.

\subsection{Graph Convolutional Networks}

Graph Convolutional Networks \cite[GCNs,][]{kipf2016semi,diego2017semantic} compute context-sensitive embeddings with respect to a graph structure.
GCNs are composed of several layers where each layer updates vertices representations based on the current representations of their neighbors.
In this work, we fed the GCN with word embeddings and a tree sample $\mT$.
For each word $x_i$, a GCN layer produces a new representation relying both on 
word embedding of $x_i$ and on embeddings of its heads and modifiers in $\mT$.
Multiple GCN layers can be stacked on top of each other.
Therefore, a vertex representation in a GCN with $k$ layers is influenced by all vertices at a maximum distance of $k$ in the graph.
Our GCN is sensitive to arc direction.

More formally, let $\mE^0 = \ve_0 \odot \dots \odot \ve_n$, where $\odot$ is the column-wise concatenation operator, be the input matrix with each column corresponding to a word in the sentence.
At each GCN layer $t$, we compute:
\begin{align*}
    \mE^{t+1} = \sigma \left(
        f(\mE^t)
        + g(\mE^t) \mT
        + h(\mE^t) \mT^\top
    \right),
\end{align*}
where $\sigma$ is an activation function, e.g. $\relu$.
Functions $f()$, $g()$ and $h()$ are distinct multi-layer perceptrons encoding different types of relationships:
self-connection, head and modifier, respectively (hyperparameters are provided in Appendix~\ref{app:nn}).
Note that each GCN layer is easily parallelizable on GPU both over vertices and over batches, either with latent or predefined structures.
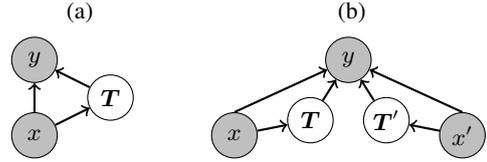
\begin{figure}[t]
    \hfill{}
    \subfloat[]{
        \begin{tikzpicture}

\node[
    circle,draw=black,
    fill=lightgray,
    minimum size=0.6cm,
    inner sep=0pt
] (x) {\small $x$};

\node[
    circle,draw=black,
    above of=x,
    yshift=-0.5cm,
    xshift=1cm,
    minimum size=0.6cm,
    inner sep=0pt
] (t) {\small $\mT$};

\node[
    circle,draw=black,
    fill=lightgray,
    above of=t,
    yshift=-0.5cm,
    xshift=-1cm,
    minimum size=0.6cm,
    inner sep=0pt
] (y) {\small $y$};

\path[->,thick]
    (x) edge (t)
    (x) edge (y)
    (t) edge (y)
;

\end{tikzpicture}
        \label{fig:network:sst}
    }
    \hfill{}
    \subfloat[]{
        \begin{tikzpicture}

\node[
    circle,
    draw=black,
    fill=lightgray,
    minimum size=0.6cm,
    inner sep=0pt
] (x1) {\small $x$};

\node[
    circle,
    draw=black,
    above of=x1,
    yshift=-0.8cm,
    xshift=1cm,
    minimum size=0.6cm,
    inner sep=0pt
] (t1) {\small $\mT$};

\node[
    circle,
    draw=black,
    right of=x1,
    xshift=2cm,
    fill=lightgray,
    minimum size=0.6cm,
    inner sep=0pt
] (x2) {\small $x'$};

\node[
    circle,
    draw=black,
    above of=x2,
    yshift=-0.8cm,
    xshift=-1cm,
    minimum size=0.6cm,
    inner sep=0pt
] (t2) {\small $\mT'$};

\node[
    circle,
    draw=black,
    fill=lightgray,
    above of=t,
    yshift=-0.5cm,
    xshift=0.5cm,
    minimum size=0.6cm,
    inner sep=0pt
] (y) {\small $y$};

\path[->,thick]
    (x1) edge (t1)
    (t1) edge (y)
    (x1.north) edge (y)
    (x2) edge (t2)
    (t2) edge (y)
    (x2.north) edge (y)
;

\end{tikzpicture}
        \label{fig:network:snli}
    }
    \hfill{}
    \caption{
        The two directed graphical models used in this work. Shaded and unshaded nodes
        represent observable and unobservable variables, respectively.
        {\bfseries \protect\subref{fig:network:sst}}
        In the sentence classification task, the output $y$ is conditioned on the input and the latent tree.
        {\bfseries \protect\subref{fig:network:snli}}
        In the natural language inference task, the output is conditioned on two sentences and their respective latent trees.
    }
    \label{fig:network}
\end{figure}

\section{Structured Latent Variable Models}

In the previous section, we explained how a dependency tree is produced for a given sentence and how we extract features from this tree with a GCN.
In our model, we assume that we do not have access to gold-standard trees and that we want to induce the best structure for the downstream task.
To this end, we introduce a probability model where the dependency structure is a latent variable (Section~\ref{sec:bayes}).
The distribution over dependency trees must be inferred from the data (Section~\ref{sec:estimation}).
This requires marginalization over dependency trees during training, which is intractable due to the large search space.%
\footnote{%
This marginalization is a sum of the network outputs over all possible projective dependency trees.
We cannot rely on the usual dynamic programming approach because we do not make any factorization assumptions in the GCN.
}
Instead, we rely on Monte-Carlo (MC) estimation.

\subsection{Graphical Model}
\label{sec:bayes}

Let $x$ be the input sentence, $y$ be the output (e.g. sentiment labelling) and $\mathcal{T}(x)$ be the set of latent structures compatible with input $x$.
We construct a directed graphical model where $x$ and $y$ are observable variables, i.e.\ their values are known during training.
However, we assume that the probability of the output $y$ is conditioned on a latent tree $\mT \in \mathcal{T}(x)$, a variable that is not observed during training:
it must be inferred from the data.
Formally, the model is defined as follows:
\begin{align}
	p_\theta(y | x)
	&= \E_{p_\theta(\mT | x)}[ p(y | x, \mT) ] \label{eq:latent_model} \\
	&= \sum_{\mT \in \mathcal{T}(x)} p_\theta(\mT | x) \times p_\theta(y | x, \mT), \nonumber
\end{align}
where $\theta$ denotes all the parameters of the model.
An illustration of the network is given in Figure~\ref{fig:network:sst}.

\subsection{Parameter Estimation}
\label{sec:estimation}

Our probability distributions are parameterized by neural networks.
Their parameters $\theta$ are learned via gradient-based optimization to maximize the log-likelihood of (observed) training data.
Unfortunately, estimating the log-likelihood of observation requires computing the expectation in Equation~\ref{eq:latent_model}, which involves an intractable sum over all valid dependency trees.
Therefore, we propose to optimize a lower bound on the log-likelihood, derived by application of  Jensen's inequality which can be efficiently estimated with the Monte-Carlo (MC) method:
\begin{align}
     \log & \;  p_{\theta}(y^i | x^i) 
=  \log \E_{\mT \sim p_{\theta}(\mT | x^i)}[ p_{\theta}(y^i | \mT, x^i) ]  \nonumber \\
    &\geq  \E_{\mT \sim p_{\theta}(\mT | x^i)}[ \log   p_{\theta}(y^i | \mT, x^i) ]. \label{eq:bound}
\end{align}
However, MC estimation introduces a non-differentiable sampling function $\mT \sim p_\theta(\mT | x^i)$  in the gradient path.
Score function estimators have been introduced to bypass this issue but suffer from high variance \cite{williams1987reinforce,fu2006gradient,schulman2015gradient}.
Instead, we propose to reparametrize the sampling process~\cite{kingma2013vae}, making it independent of the learned parameter $\theta$~: in such case, the sampling function is outside of the gradient path.
To this end, we rely on the Perturb-and-MAP framework \cite{papandreou2011perturb}.
Specifically, we perturb the potentials (arc weights) with samples from the Gumbel distribution and compute the most probable structure with the perturbed potentials:
\begin{align}
    \emG_{h, m} &\sim \mathcal{G}(0, 1), \\
    \widetilde{\mW} &= \mW + \mG, \\
    \mT &= \argmax_{\mT \in \mathcal{T}(x)} \sum_{h, m} \emT_{h, m} \widetilde{\emW}_{h, m}. \label{eq:sampling:argmax}
\end{align}
Each element of the matrix $\mG \in \R^{n \times n}$ contains random samples from the Gumbel distribution%
\footnote{That is $\emG_{h, m} = -\log(-\log(\emU_{h, m}))$ where $\emU_{h, m}$ is sampled from the uniform distribution on the interval $(0, 1)$.}
which is independent from the network parameters $\theta$, hence there is no need to backpropagate through this path in the computation graph. Note that, unlike the  Gumbel-Max trick~\cite{maddison2014sampling}, sampling with Perturb-and-MAP is approximate, as the noise is factorizable: we add noise to individual arc weights rather than to scores of entire trees (which would not be tractable). This is the first source of bias in our gradient estimator.
The maximization in Equation~\ref{eq:sampling:argmax} can be computed using the algorithm of \citet{eisner1996dep}.
We stress that the marginalization in Equation~\ref{eq:latent_model} and MC estimated sum over trees capture high-order statistics, which is fundamentally different from computing edge marginals, i.e.\ structured attention~\cite{kim2017structuredatt}.
Unfortunately, the estimated gradient of the reparameterized distribution over parse trees is ill-defined (either undefined or null).
We tackle this issue in the following section.
\begin{algorithm}[t]
    \begin{algorithmic}[1]
\Function{build-uright}{$i, j, \widetilde{\mW}$}
    \State{$\vs \gets$~null-initialized vec. of size $j-i$}
    \For{$i \leq k < j$}
        \State{\hspace{-0.8em} 
            \begin{varwidth}[t]{\linewidth}
                $\evs_{i-k}$~$\gets$~$[i, k, \rightarrow, \top]+[k+1, j, \leftarrow, \top]$ \par
            \end{varwidth}
        }
    \EndFor
    \State{$\vb \gets $ \Call{one-hot-argmax}{$\vs$}} \label{alg:argmax}
    \State{$\textsc{backptr}[i, j, \rightarrow, \bot]$~$\gets$~$\vb$}
    \State{$\textsc{weight}[i, j, \rightarrow, \bot]$~$\gets$~$\vb^\top \vs + \emW_{j, i}$}
\EndFunction
\end{algorithmic}
     \caption{
        This function computes the chart values for items of the form $[i, j, \rightarrow, \bot]$ by searching the set of antecedents that maximizes its score.
        Because these items assume a dependency from $x_i$ to $x_j$, we add $\emW_{i, h}$ to the score.
     }
     \label{alg:forward}
\end{algorithm}

\begin{algorithm}[t]
    \begin{algorithmic}[1]
  \Function{Backtrack-uright}{$i, j, \mT$}
      \State{$\emT_{i,j} \gets \textsc{contrib}[i, j, \rightarrow, \bot]$}
      \State{$\vb \gets \textsc{backptr}[i, j, \rightarrow, \bot]$}
      \For{$i \leq k < j$}
          \State{$\textsc{contrib}[i,k, \rightarrow, \top] \overset{+}{\gets} \evb_{i-k} \emT_{i,j}$}
          \State{$\textsc{contrib}[k+1, j, \leftarrow, \top] \overset{+}{\gets} \evb_{i-k} \emT_{i,j}$}
      \EndFor
  \EndFunction
\end{algorithmic}
     \caption{
    	If item $[i, j, \rightarrow, \bot]$ has contributed the optimal objective,
        this function sets $\emT_{i, j}$ to $1$.
        Then, it propagates the contribution information to its antecedents.
     }
     \label{alg:backtrack}
\end{algorithm}

\section{Differentiable Dynamic Programming}
\label{sect:diff-dp}
Neural networks parameters are learned using (variants of) the stochastic gradient descent algorithm.
The gradient is computed using the back-propagation algorithm that rely on partial derivative of each atomic operation in the network.%
\footnote{%
    There are some exception where a sub-derivative is enough, for example for the $\relu$ non-linearity.
}
The perturb-and-MAP sampling process relies on the dependency parser (Equation~\ref{eq:sampling:argmax}) which contains ill-defined derivatives.
This is due to the usage of constrained $\argmax$ operations \cite{gould2016differentiating,mensch2018differentiable} in the algorithm of \citet{eisner1996dep}.
Let $\mathcal{L}$ be the training loss, backpropagation is problematic because of the following operation:
\begin{align*}
    \frac{\partial \mathcal{L}}{\partial \widetilde{\mW}}
    =
    \frac{\partial \mathcal{L}}{\partial \mT}
    \frac{\partial \mT}{\partial \widetilde{\mW}}
\end{align*}
where $\frac{\partial \mT}{\partial \widetilde{\mW}}$ is the partial derivative with respect to the dependency parser (Equation~\ref{eq:sampling:argmax}) which is null almost everywhere,
i.e.\ there is no descent direction information.
We follow previous work and use a differentiable dynamic programming surrogate \cite{mensch2018differentiable,corro2018differentiable}. The use of the surrogate is the second source of bias in our gradient estimation.

\subsection{Parsing with Dynamic Programming}

The projective dependency parser of \citet{eisner1996dep} is a dynamic program that recursively builds a chart of items representing larger and larger spans of the input sentence.
Items are of the form $[i, j, d, c]$ where:
$0 \leq i \leq j \leq n$ are the boundaries of the span;
$d \in \{\rightarrow, \leftarrow\}$ is the direction of the span, i.e.\ a right span $\rightarrow$ (resp. left span $\leftarrow$) means that all the words in the span are descendants of $x_i$ (resp. $x_j$) in the dependency tree;
$c \in \{\top, \bot \}$ indicates if the span is complete ($\top$) or incomplete ($\bot$) in its direction.
In a complete right span, $x_j$ cannot have any modifiers on its right side.
In a complete left span, $x_i$ cannot have any modifier on its left side.
A set of deduction rules defines how the items can be deduced from their antecedents.

The algorithm consists of two steps.
In the first step, items are deduced in a bottom-up fashion and the following information is stored in the chart: the maximum weight that can be obtained by each item and backpointers to the antecedents that lead to this maximum weight (Algorithm~\ref{alg:forward}).
In the second step, the backpointers are used to retrieve the items corresponding to the maximum score and values in $\mT$ are set accordingly (Algorithm~\ref{alg:backtrack}).%
\footnote{%
The second step is often optimized to have linear time complexity instead of cubic.
Unfortunately, this change is not compatible with the continuous relaxation we propose.
}

\subsection{Continuous Relaxation}

The $\onehotargmax$ operation on line \ref{alg:argmax} in Algorithm~\ref{alg:forward} can be written as follows:
\begin{align*}
    &\argmax_{\vb \geq 0} \quad \sum_k \evb_k \evs_k && \text{s.t.} \sum_k \evb_k = 1. \\
\intertext{
    It is known that a continuous relaxation of $\argmax$ in the presence of inequality constraints can be obtained by introducing a penalizer that prevents activation of inequalities at the optimal solutions \cite{gould2016differentiating}:
}
    &\argmax_{\vb \geq 0} \quad \sum_k \evb_k \evs_k - \Omega(\vb) && \text{s.t.} \sum_k \evb_k = 1.
\end{align*}
Several $\Omega$ functions have been studied in the literature for different purposes, including logarithmic and inverse barriers for the interior point method \cite{den1994inverse,potra2000interior} and negative entropy for deterministic annealing \cite{rangarajan2000annealing}.
When using negative entropy, i.e.\ $\Omega(\vb) = \sum_k \evb_k \log \evb_k$, solving the penalized $\onehotargmax$ has a closed form solution that can be computed using the $\softmax$ function \cite{boyd2004convex}, that is:
\begin{align*}
    \evb_k = \frac{\exp(\evs_k)}{\sum_{k'} \exp(\evs_{k'})}.
\end{align*}

Therefore, we replace the non-differentiable $\onehotargmax$ operation in Algorithm~\ref{alg:forward} with a softmax in order to build a smooth and fully differentiable surrogate of the parsing algorithm.s
\section{Controlled Experiment}
\label{sec:exp:listops}

We first experiment on a toy task. 
The task is designed in such a way that there exists a simple projective dependency grammar which turns it into a trivial problem.
We can therefore perform  thorough analysis of the latent tree induction method.

\subsection{Dataset and Task}

The ListOps dataset \cite{nangia2018listops} has been built specifically to test structured latent variable models.
The task is to compute the result of a mathematical expression written in prefix notation.
It has been shown easy for a Tree-LSTM that follows the gold underlying structure but  most latent variable models fail to induce it.
Unfortunately, the task is not compatible with our neural network because it requires  propagation of information from the leafs to the root node,
which is not possible for a GCN with a fixed number of layers.
Instead, we transform the computation problem into a tagging problem:
the task is to tag the valency of operations, i.e.\ the number of operands they have.

\begin{figure}[t]
    \centering
    \input{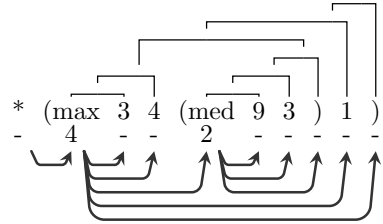}
    \caption{
        An example from the ListOps dataset.
        Numbers below operation tokens are valencies.
        {\bfseries (top)} the original unlabelled phrase-structure.
        {\bfseries (bottom)} our dependency conversion: each dependency represents either an operand to argument relation or a closing parenthesis relation.
    }        
    \label{fig:listops}
\end{figure}

We transform the original unlabelled binary phrase-structure into a dependency structure by following a simple head-percolation table:
the head of a phrase is always the head of its left argument.
The resulting dependencies represent two kinds of relation: operand to argument and operand to closing parenthesis (Figure~\ref{fig:listops}).
Therefore, this task is trivial for a GCN trained with gold dependencies: it simply needs to count the number of outgoing arcs minus one (for operation nodes).
In practice, we observe 100\% tagging accuracy with the gold dependencies.
 
\subsection{Neural Parametrization}

We build a simple network where a BiLSTM is followed by deep dotted attention which computes the dependency weights (see Equation~\ref{eq:parse-pred}). In these experiments, unlike Section~\ref{sect:real-experiments}, GCN does not have access to input tokens (or corresponding BiLSTM states): it is fed `unlexicalized' embeddings (i.e.\ the same vector is used as input for every token).%
\footnote{%
To put it clearly, we have two sets of learned embeddings:
a set of lexicalized embeddings used for the input of the BiLSTM and a single unlexicalized embedding used for the input of the GCN.
}
Therefore, the GCN is forced to rely on tree information alone (see App.~\ref{app:listops} for hyperparameters).

There are several ways to train the neural network.
First, we test the impact of MC estimation at training.
Second, we choose when to use the continuous relaxation.
One option is to use a {\it Straight-Through} estimator \cite[ST, ][]{bengio2013estimating,jang2017categorical}:
during the forward pass, we use a discrete structure as input of the GCN, but during the backward pass we use the differentiable surrogate to compute the partial derivatives.
Another option is to use the differentiable surrogate for both passes ({\it Forward relaxed}).
 As our goal here is to study
induced discrete structures, we do not use relaxations at test time.
We compare our model with the non-stochastic version, i.e.\ we set $\mG = 0$.

\subsection{Results}

The attachment scores and the tagging accuracy are provided in Table~\ref{tab:listops}.
We draw two 
conclusions from these results.
First, using the ST estimator hurts performance, even though we do not relax at test time.
Second, the MC approximations, unlike the non-stochastic model, produces latent structures
almost identical to gold trees. 
The non-stochastic version is however relatively successful in terms of tagging accuracy: we hypothesize that the LSTM model solved the problem and uses trees as messages to communicate solutions. 
See extra analysis in App.~\ref{app:listops:training}.%
\footnote{
    This results are not cherry-picked to favor the MC model.
    We observed a deviation of $\pm0.54\%$ in attachment score  for the non-stochastic model,
    whereas, for MC sampling, all except one achieved an attachment score above $99.7$ (out of 5 runs).
}

\begin{table}[t]
    \centering
    
\begin{tabular}{l|c|c}
     \multicolumn{1}{l}{} & {\bfseries Acc.} & {\bfseries Att.} \\
     \hline
     \multicolumn{3}{l}{\bfseries Latent tree - $\mG = 0$} \\
     \hline
     Forward relaxed & 98.1	& 83.2 \\
     Straight-Through & 70.8	& 33.9 \\
     \hline
     \multicolumn{3}{l}{\bfseries Latent tree - MC training} \\
     \hline
     Forward relaxed & {\bfseries 99.6} & {\bfseries 99.7} \\
     Straight-Through & 77.0	& 83.2 \\
\end{tabular}

    \caption{
        ListOps results: tagging accuracy (Acc.) and attachment score for the latent tree grammar (Att.).
    }
    \label{tab:listops}
\end{table}

\begin{table}[t]
    \centering

\begin{tabular}[t]{l|c|c}   
    & {\bfseries Acc.} & {\bfseries \#Params} \\
    \hline
    \multicolumn{3}{l}{\bfseries \citet{yogatama2016reinforce}} \\
    \hline
    *100D SPINN & 80.5 & 2.3M \\
    \hline
    \multicolumn{3}{l}{\bfseries \citet{maillard2017jointly}} \\
    \hline
    LSTM & 81.2 & 161K \\
    *Latent Tree-LSTM & 81.6 & 231K \\
    \hline
    \multicolumn{2}{l}{\bfseries \citet{kim2017structuredatt}} \\
    \hline
    No Intra Attention & 85.8 & - \\
    Simple Simple Att. & 86.2 & - \\
    *Structured Attention & 86.8 & - \\
    \hline
    \multicolumn{3}{l}{\bfseries \citet{choi2018gumbeltree}} \\
    \hline
    *100D ST Gumbel Tree & 82.6 & 262K \\
    *300D ST Gumbel Tree & 85.6 & 2.9M \\
    *600D ST Gumbel Tree & 86.0 & 10.3M \\
    \hline
    \multicolumn{3}{l}{\bfseries \citet{niculae2018dyncg}} \\
    \hline
    Left-to-right Trees & 81.0 & - \\
    Flat & 81.7 & - \\
    Treebank & 81.7 & - \\
    *SparseMAP & 81.9 & - \\
    \hline
    \multicolumn{3}{l}{\bfseries \citet{liu2018structuredatt}} \\
    \hline
    175D No Attention & 85.3 & 600K \\
    *100D Projective Att. & 86.8 & 1.2M \\
    *175D Non-projective Att. & 86.9 & 1.1M \\
    \hline
    \multicolumn{3}{l}{\bfseries This work} \\
    \hline
    No Intra Attention & 84.4 & 382K \\  
    Simple Intra Att. & 83.8 & 582K \\ 
    *Latent Tree + 1 GCN & 85.2 & 703K \\ 
    *Latent Tree + 2 GCN & 86.2 & 1M \\ 
    \hline
\end{tabular}
    \caption{
        SNLI results and number of network parameters (discarding word embeddings).
        Stars indicate latent tree models.
    }
    \label{tab:snli:results}
\end{table}

\begin{table*}[]
    \centering
    \subfloat[]{
        %
%

\begin{tabular}[t]{l|c}
    \multicolumn{2}{c}{} \\
    \hline
    \multicolumn{2}{l}{\bfseries \citet{socher2013sst}} \\
    \hline
    Bigram & 83.1 \\
    Naive Bayes & \\
    \hline
    \multicolumn{2}{l}{\bfseries \citet{niculae2018dyncg}} \\
    \hline
    CoreNLP & 83.2 \\
    *Latent tree & 84.7 \\
    \hline
    \multicolumn{2}{l}{\bfseries  This work} \\
    \hline
    CoreNLP & 83.8 \\
    *Latent tree & 84.6 \\
    \hline
\end{tabular}
        \label{tab:sst:results}
    }
    \subfloat[]{
        %
%

\begin{tabular}[t]{l|c}   
    & {\bfseries Acc.} \\
    \hline
    \multicolumn{2}{l}{\bfseries \citet{williams2018latent}} \\
    \hline 
    300D LSTM & 69.1 \\
    *300D SPINN & 66.9 \\
    300D Balanced Trees & 68.2 \\
    *300D ST Gumbel Tree & 69.5 \\
    *300D RL-SPINN & 67.3 \\
    \hline
    \multicolumn{2}{l}{\bfseries This work} \\
    \hline 
    No Intra Attention & 68.1 \\ 
    *Latent tree + 1 GCN & 71.5 \\ 
    *Latent tree + 2 GCN & 73.0 \\
    \hline
\end{tabular}
        \label{tab:multinli:results}
    }
    \subfloat[]{
        %
%

\begin{tabular}{l|c|c}
    & {\bfseries Match} & {\bfseries Mis.} \\
    \hline
    \multicolumn{3}{l}{\bfseries Baselines} \\
    \hline 
    No Intra Att & 68.5 & 68.9 \\ 
    Simple Intra Att & 67.9 & 68.4 \\ 
    \hline 
    \multicolumn{3}{l}{\bfseries Left-to-right trees} \\
    \hline 
    
    1 GCN & 71.2 & 71.8 \\
    2 GCN & 72.3 & 71.1 \\
    
    \hline
    \multicolumn{3}{l}{\bfseries Latent head selection model} \\
    \hline 
    1 GCN & 69.0 & 69.4 \\ 
    2 GCN & 68.7 & 69.6 \\
    \hline
    \multicolumn{3}{l}{\bfseries Latent tree model} \\ 
    \hline 
    1 GCN & 71.9 & 71.7 \\ 
    2 GCN & 73.2 & 72.9 \\
    \hline
\end{tabular}

        \label{tab:ablation}
    }
    \caption{
        {\bfseries \protect\subref{tab:sst:results}}
        SST results.
        Stars indicate latent tree models.
        {\bfseries \protect\subref{tab:multinli:results}}
        MultiNLI results.
        Stars indicate latent tree models.
        {\bfseries \protect\subref{tab:ablation}}
        Ablation tests on MultiNLI (results on the matched and mismatched development sets).
    }
\end{table*}

\section{Real-world Experiments}
\label{sect:real-experiments}

We evaluate our method on two real-world problems: a sentence comparison task (natural language inference, see Section~\ref{sec:nli}) and a sentence classification problem (sentiment classification, see Section~\ref{sec:st}). Besides using the differentiable dynamic programming method, our approach also differs from previous work in that we use GCNs followed by a pooling operation, whereas most previous work used Tree-LSTMs.
Unlike Tree-LSTMs, GCNs are trivial to parallelize over batches on GPU.%

\subsection{Natural Language Inference}
\label{sec:nli}

The Natural Language Inference (NLI) problem is a  task developed to test sentence understanding capacity.
Given a premise sentence and a hypothesis sentence, the goal is to predict a relation between them: entailment, neutral or contradiction.
We evaluate on the Stanford NLI (SNLI) and the Multi-genre NLI (MultiNLI) datasets.
Our network is based on the decomposable attention (DA) model of \citet{parikh2016decomposable}.
We induce structure of both the premise and the hypothesis (see Equation~\ref{eq:parse-pred} and Figure~\ref{fig:network:snli}).
Then, we run a GCN over the tree structures followed by inter-sentence attention.
Finally, we apply max-pooling  for each sentence and feed both sentence embeddings into a MLP to predict the label. Intuitively, using GCNs yields a form of 
intra-attention.
See the hyper-parameters  in Appendix~\ref{app:nn:nli}.

\textbf{SNLI:}
The dataset contains almost 0.5m training instances extracted from image captions \cite{bowman2015snli}.
We report results in Table~\ref{tab:snli:results}.

Our model outperforms both no intra-attention and simple intra-attention baselines\footnote{The attention weights are computed in the same way as scores for tree prediction, i.e. using Equation~\ref{eq:parse-pred}.}
with 1 layer of GCN ($+0.8$) or two layers ($+1.8$). The improvements with using multiple GCN hops, here and on MultiNLI (Table~\ref{tab:multinli:results}), suggest that higher-order information is beneficial.%
\footnote{
    In contrast, multiple hops relying on edge marginals was not beneficial~\cite{liu2018structuredatt}, personal communication.
}
It is hard to compare different tree induction methods as they build on top of different baselines, however, it is clear that our model delivers results 
comparable with most accurate tree induction methods \cite{kim2017structuredatt,liu2018structuredatt}. The improvements from using latent structure exceed these reported in previous work.


\textbf{MultiNLI:}
MultiNLI is a broad-coverage NLI corpus
\citet{williams2018multinli}: 
the sentence pairs originate from 5 different genres of written and spoken English.
This dataset is particularly interesting because sentences are longer than in SNLI, making it more challenging for baseline models.%
\footnote{
    The average sentence length in SNLI (resp. MultiNLI) is $11.16$ (resp. $16.79$).
    There is $21\%$ (resp. $42\%$) of sentence longer than 15 words in SNLI (resp. MultiNLI).
}
We follow the evaluation setting in \citet{williams2018multinli,williams2018latent}: 
we include the SNLI training data, use the matched development set for early stopping and evaluate on the matched test set.
We use the same network and parameters as for SNLI.
We report results in Table~\ref{tab:multinli:results}.

The DA baseline (`No Intra Attention') performs slightly better ($+0.6\%$) than the original BiLSTM baseline.
Our latent tree model significantly improves over our the baseline, either with a single layer GCN ($+3.4$\%) or with a 2-layer GCN ($+4.9\%$).
We observe a larger gap than on SNLI, which is expected given that MultiNLI is more complex. We perform extra ablation tests on MultiNLI in Section~\ref{sec:analysis}.

\begin{figure*}
    \vspace{-1ex}
    \centering
    \includegraphics{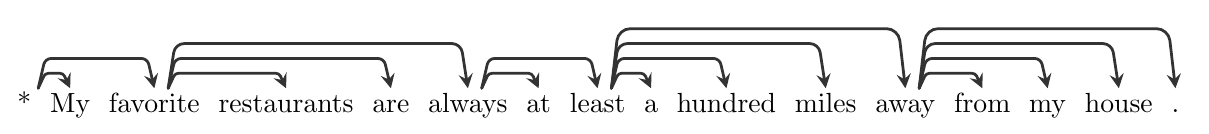}
    \begin{minipage}[b]{0.4\textwidth}
    \includegraphics{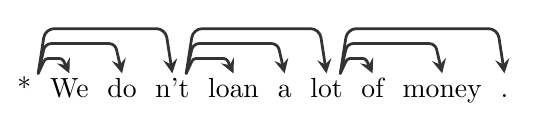}
    \end{minipage}%
    \begin{minipage}[b]{0.6\textwidth}
    \includegraphics{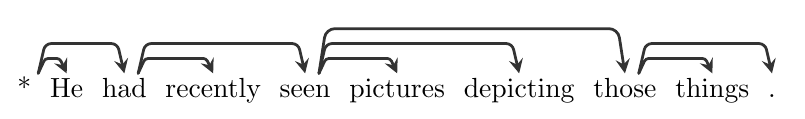}
    \vspace{-1ex}
    \end{minipage}
    \caption{
         Examples of  trees induced on the matched development set of  MultiNLI, the model using  2 GCN layers.
    }
    \label{tab:examples}
\end{figure*}

\subsection{Sentiment Classification}
\label{sec:st}

We experiment on the Stanford Sentiment Classification dataset \cite{socher2013sst}.
The original dataset contains predicted constituency structure with manual sentiment labeling for each phrase.
By definition, latent tree models cannot use the internal phrase annotation.
We follow the setting of \citet{niculae2018dyncg} and compare to them
in two set-ups:
(1) with syntactic dependency trees predicted by CoreNLP \cite{manning2014corenlp};
(2) with latent dependency trees.
Results are reported in Table~\ref{tab:sst:results}.

First, we observe that  the bag of bigrams baseline of \citet{socher2013sst} achieves results comparable to all structured models. This suggest that the dataset may not be well suited for evaluating structure induction methods. 
Our latent dependency model slighty improves ($+0.8$) over the CoreNLP baseline.
However, we observe that while our baseline is better than the one of \citet{niculae2018dyncg}, their latent tree model slightly outperforms ours ($+0.1$).
We hypothesize that graph convolutions may not be optimal for this task.

\subsection{Analysis}
\label{sec:analysis}

\textbf{(Ablations)} 
In order to test if the tree constraint is important, we do ablations on MultiNLI  with two models:
one with a latent projective tree variable (i.e. our full model) and one with a latent head selection model that does not impose any constraints on the structure. 
The estimation approach and the model
are identical, except for the lack of the tree constraint (and hence dynamic programming) in the ablated model.
We report results 
on development sets in Table~\ref{tab:ablation}.
We observe that the latent tree models
outperform the alternatives.

Previous work \cite[e.g.,][]{niculae2018dyncg} included comparison with balanced trees, flat trees and left-to-right (or right-to-left) chains. Flat trees are pointless with the GCN + DA combination: the corresponding pooling operation is already done in DA.  Though balanced trees are natural with bottom-up computation of TreeLSTMs, for GCNs they would result in embedding
essentially random subsets of words.  Consequently, we compare only to left-to-right chains of dependencies.\footnote{They are the same as right-to-left ones, as our GCNs treat both directions equivalently.}
This approach is substantially less accurate than our methods, especially for out-of-domain (i.e. mismatched) data.

\noindent
\textbf{(Grammar)} We also investigate the structure of the induced grammar.
We report the latent structure of three sentences in Figure~\ref{tab:examples}.
We observe that sentences are divided into spans, where each span is represented with a series of left dependencies. Surprisingly, the model chooses to use only left-to-right dependencies.
The neural network does not include a RNN layer, so this may suggest that the grammar is trying to reproduce an recurrent model while also segmenting the sentence in phrases.

\noindent
\textbf{(Speed)}
We use  a $\mathcal{O}(n^3)$-time parsing algorithm. 
Nevertheless, our model is  efficient: one epoch on SNLI takes 470 seconds, only
140 seconds longer than with the $\mathcal{O}(n^2)$-time latent-head version of our model (roughly equivalent to classic self-attention). The latter model is computed  on GPU (Titan X) while ours uses CPU (Xeon E5-2620) for the dynamic program and GPU for running the rest of the network.

\section{Related work}

Recently, there has been growing interest in providing an inductive bias in neural network by forcing layers to represent 
 tree structures \cite{kim2017structuredatt,maillard2017jointly,choi2018gumbeltree,niculae2018dyncg,williams2018latent,liu2018structuredatt}.
\citet{maillard2017jointly} also operates on a chart but, rather than modeling discrete trees, uses a soft-gating approach to mix representations of constituents in each given cell.
While these models showed consistent improvement over comparable baselines, they do not seem to explicitly capture syntactic or semantic structures \cite{williams2018latent}.
\citet{nangia2018listops} introduced the ListOps task where the latent structure is essential to predict correctly the downstream prediction.
Surprisingly, the models of \citet{williams2018latent} and \citet{choi2018gumbeltree} failed.
Much recent work in this context relies on latent variables, though we are not aware of any work closely related to ours.
Differentiable structured layers in neural networks have  been explored for semi-supervised parsing,
for example by learning an auxiliary task on unlabelled data \cite{head2018sigpot} or using a variational autoencoder~\cite{corro2018differentiable}.

Besides research focused on inducing task-specific structures, 
another line of work, grammar induction, focused on
unsupervised induction of linguistic structures.
These methods typically rely on unlabeled texts and are evaluated by comparing the induced structures to actual syntactic annotation \cite{klein2005unsupervised,shen2017jointsyntlex,htut2018replication}.

\section{Conclusions}
We introduced a novel approach to latent tree learning: a relaxed version of stochastic differentiable dynamic programming which allows for efficient sampling of  projective dependency trees and enables end-to-end differentiation.
We demonstrate effectiveness of our approach on both synthetic and real tasks.
The analyses confirm importance of the tree constraint.
Future work will investigate constituency structures and new neural architectures for latent structure incorporation.

\section*{Acknowledgments}
We thank Maximin Coavoux and Serhii Havrylov for their comments and suggestions.
We are grateful to Vlad Niculae for the help with pre-processing the SST data.
We also thank the  anonymous  reviewers  for  their  comments.
The project was supported by the Dutch National Science Foundation (NWO VIDI 639.022.518) and European Research Council (ERC Starting Grant BroadSem 678254).

\bibliography{bib}
\bibliographystyle{acl_natbib}

\clearpage
\appendix
\section{Neural Parametrization}
\label{app:nn}

\paragraph{(Implementation)}
We implemented our neural networks with the C++ API of the Dynet library \cite{dynet}.
The continuous relaxation of the parsing algorithm is implemented as a custom computation node.

\paragraph{(Training)}
All networks are trained with Adam initialized with a learning rate of $0.0001$ and batches of size 64.
If the dev score did not improve in the last 5 iterations, we multiply the learning rate by $0.9$ and load the best known model on dev.
For the ListOps task, we run a maximum of 100 epochs, with exactly 100 updates per epoch.
For NLI and SST tasks, we run a maximum of 200 epochs, with exactly 8500 and 100 updates per epoch, respectively.

All MLPs and GCNs have a dropout ratio of $0.2$ except for the ListOps task where there is no dropout.
We clip the gradient if its norm exceed $5$.

\subsection{ListOps Valency Tagging}
\label{app:listops}

\paragraph{(Dependency Parser)}
Embeddings are of size 100.
The BiLSTM is composed of two stacks (i.e. we first run a left-to-right and a right-to-left LSTM, then we concatenate their outputs and finally run a left-to-right and a right-to-left LSTM again) with one single hidden layer of size 100.
The initial state of the LSTMs are fixed to zero.

The MLPs of the dotted attention have 2 layers of size 100 and a $\relu$ activation function

\paragraph{(Tagger)}
The unique embedding is of size 100.
The GCN has a single layer of size 100 and a $\relu$ activation.
Then, the tagger is composed of a MLP with a layer of size 100 and a $\relu$ activation followed by a linear projection into the output space (i.e.\ no bias, no non-linearity).

\subsection{Natural Language Inference}
\label{app:nn:nli}

All activation functions are $\relu$.
The inter-attention part and the classifier are exactly the same than in the model of \citet{parikh2016decomposable}.

\paragraph{(Embeddings)}
Word embeddings of size 300 are initialized with Glove and are not updated during training.
We initialize 100 unknown word embeddings where each value is sampled from the normal distribution.
Unknown words are mapped using a hashing method.

\paragraph{(GCN)}
The embeddings are first passed through a one layer MLP with an output size of 200.
The dotted attention is computed by two MLP with two layers of size 200 each.
Function $f()$, $g()$ and $h()$ in the GCN layers are one layer MLPs without activation function.
The $\sigma$ activation function of a GCN is $\relu$.
We use dense connections for the GCN.

\subsection{Sentiment Classification}
\label{app:nn:sst}

\paragraph{(Embeddings)}
We use Glove embeddings of size 300.
We learn the unknown word embeddings.
Then, we compute context sensitive embeddings with a single-stack/single-layer BiLSTM with a hidden-layer of size 100.

\paragraph{(GCN)}
The dotted attention is computed by two MLP with one layer of size 300 each.
There is no distance bias in this model.
Function $f()$, $g()$ and $h()$ in the GCN layers are one layer MLPs without activation function.
The $\sigma$ activation function of a GCN is $\relu$.
We do not use dense connections in this model.

\paragraph{(Output)}
We use a max-pooling operation on the GCN outputs followed by an single-layer MLP of size 300.

\section{Illustration of the Continuous Relaxation}
\label{app:argmax}

\begin{figure*}
    \centering
    \subfloat[]{
        \includegraphics{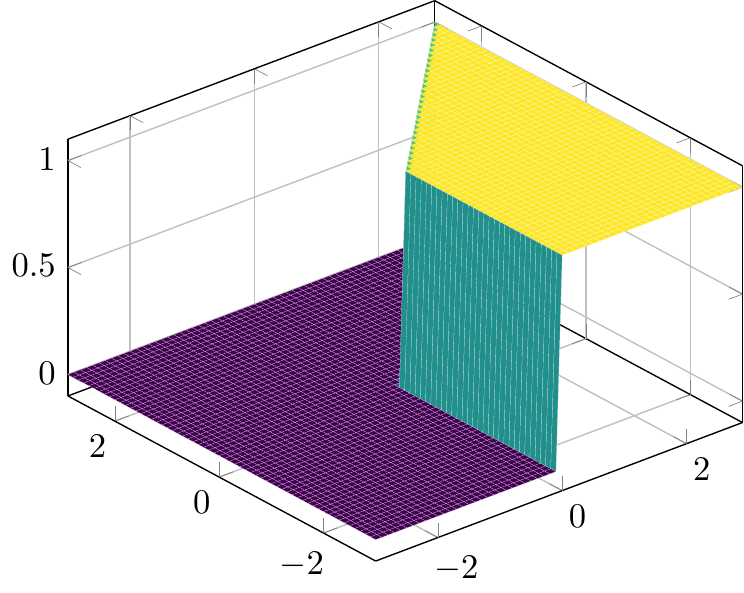}
        \label{fig:argmax:a}
    }
    \subfloat[]{
        \includegraphics{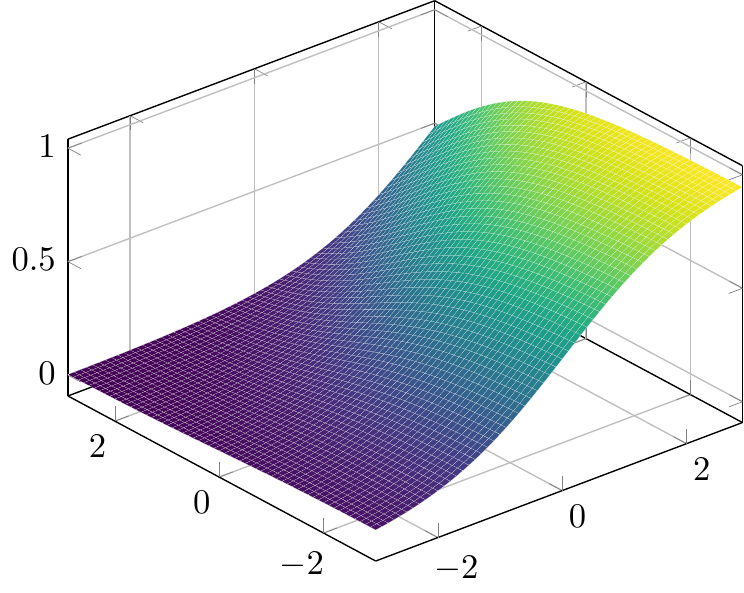}
        \label{fig:argmax:b}
    }
    \caption{
        \protect\subref{fig:argmax:a}
        Single output of an $\argmax$ function.
        The derivative is null almost everywhere, i.e.\ there is no descent direction.
        \protect\subref{fig:argmax:b}
        Single output of the differentiable relaxation.
        The derivatives are non-null.
    }
    \label{fig:argmax}
\end{figure*}

Too give an intuition of the continuous relaxation, we plot the $\argmax$ function and the penalized $\argmax$ in Figure~\ref{fig:argmax}.
We plot the first output for input $(x_1, x_2, 0)$.

\section{ListOps Training}
\label{app:listops:training}

\begin{figure*}
    \centering
    \subfloat[][Accuracy]{
        \begin{tikzpicture}[scale=0.85]
    \begin{axis}[
        xlabel={epoch},
        ylabel={accuracy},
        xmin=0, xmax=100,
        ymin=0, ymax=100,
        no marks,
        thick
    ]
        \addplot[red,solid] table[
            x expr=\coordindex,
            y index=0,
            y expr=\thisrowno{0}*100
        ]{data/nost_deterministic_dev.100};
        
        \addplot[blue,dashed] table[
            x expr=\coordindex,
            y index=0,
            y expr=\thisrowno{0}*100
        ]{data/nost_perturb_dev.100};
        
        \addplot[dotted] table[
            x expr=\coordindex,
            y index=0,
            y expr=\thisrowno{0}*100
        ]{data/st_deterministic_dev.100};
        
        \addplot[green,dashdotted] table[
            x expr=\coordindex,
            y index=0,
            y expr=\thisrowno{0}*100
        ]{data/st_perturb_dev.100};
     \end{axis}
\end{tikzpicture}
    }%
    \hfill{}%
    \subfloat[][Attachment score]{
        \begin{tikzpicture}[scale=0.85]
    \begin{axis}[
        xlabel={epoch},
        ylabel={attachment score},
        xmin=0, xmax=100,
        ymin=0, ymax=100,
        no marks,
        thick
    ]
        \addplot[red,solid] table[
            x expr=\coordindex,
            y index=0
        ]{data/nost_deterministic_uas.100};
        
        \addplot[blue,dashed] table[
            x expr=\coordindex,
            y index=0
        ]{data/nost_perturb_uas.100};
        
        \addplot[dotted] table[
            x expr=\coordindex,
            y index=0
        ]{data/st_deterministic_uas.100};
        
        \addplot[green,dashdotted] table[
            x expr=\coordindex,
            y index=0
        ]{data/st_perturb_uas.100};
     \end{axis}
\end{tikzpicture}
    }
    \caption{
        Accuracy of tagging and attachment score of the latent tree during training.
        {\bfseries (red solid line)} Non-stochastic training with forward relaxation. 
        {\bfseries (blue dashed line)} MC training with forward relaxation. 
        {\bfseries (black dotted)} Non-stochastic training with backward relaxation.
        {\bfseries (green dashdotted)} MC with backward relaxation.
    }
    \label{fig:listops:training}
\end{figure*}
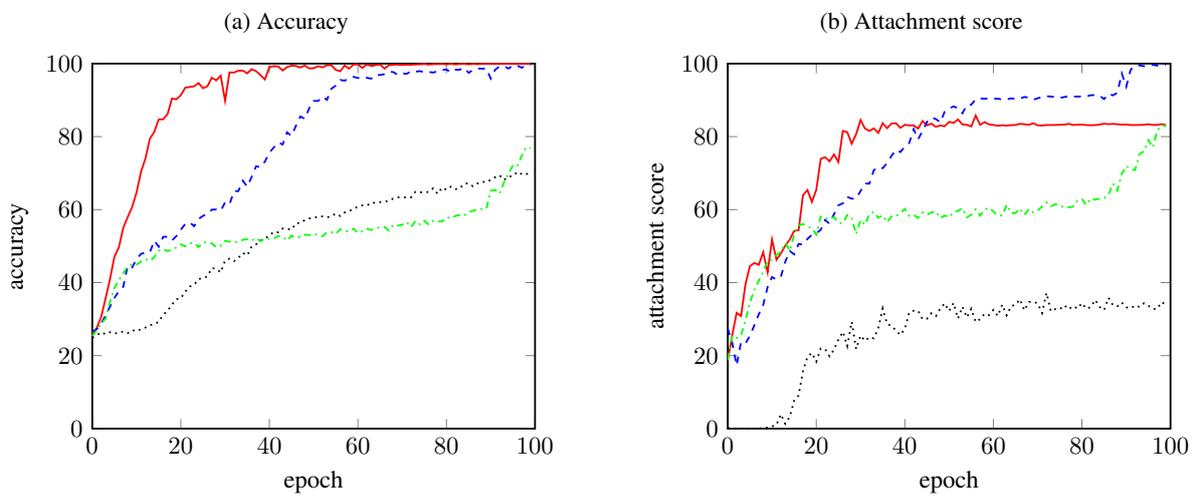

We plot tagging accuracy and attachment score with respect to the training epoch in Figure~\ref{fig:listops:training}.
On the one hand, we observe that the non-stochastic versions converges way faster in both metrics: we suspect that it develops an alternative protocol to pass information about valencies from LSTM to the GCN.
On the other hand, MC sampling may have a better exploration of the search space but it is slower to converge.

We stress that training with MC estimation results in the latent tree corresponding (almost) perfectly to the gold grammar.

\section{Fast differentiable dynamic program implementation}

In order to speed up training, we build a a fast the differentiable dynamic program (DDP) as a custom computational node in Dynet and use it in a static graph.
Instead of relying on masking, we add an input the DDP node that contains the sentence size~: therefore, even if the size of the graph is fixed,
the cubic-time algorithm is run on the true input length only.
Moreover, instead of allocating memory with the standard library functionnality, we use the fast scratch memory allocator of Dynet.

\end{document}